\documentclass[]{fairmeta}
\usepackage{amsthm,amsmath,amssymb}
\usepackage{graphicx}
\usepackage{natbib}
\usepackage{subcaption}
\usepackage{algorithm,algorithmicx,algpseudocode}
\newtheorem{theorem}{Theorem}[]

\newtheorem{remark1}[theorem]{Remark}

\title{TeleStyle:  Content-Preserving Style Transfer in Images and Videos}
\author{Shiwen Zhang}
\author{Xiaoyan Yang}
\author{Bojia Zi}
\author{Haibin Huang}
\author{Chi Zhang}
\author{Xuelong Li}
\affiliation{TeleAI}
\project {\url{https://tele-ai.github.io/TeleStyle/}}
\code{\url{https://github.com/Tele-AI/TeleStyle}}

\begin{document}

\abstract{
Content-preserving style transfer—generating stylized outputs based on content and style references—remains a significant challenge for Diffusion Transformers (DiTs) due to the inherent entanglement of content and style features in their internal representations. In this technical report, we present TeleStyle, a lightweight yet effective model for both image and video stylization. Built upon Qwen-Image-Edit, TeleStyle leverages the base model’s robust capabilities in content preservation and style customization. To facilitate effective training, we curated a high-quality dataset of distinct specific styles and further synthesized triplets using thousands of diverse, in-the-wild style categories. We introduce a Curriculum Continual Learning framework to train TeleStyle on this hybrid dataset of clean (curated) and noisy (synthetic) triplets. This approach enables the model to generalize to unseen styles without compromising precise content fidelity. Additionally, we introduce a video-to-video stylization module to enhance temporal consistency and visual quality. TeleStyle achieves state-of-the-art performance across three core evaluation metrics: style similarity, content consistency, and aesthetic quality. Code and pre-trained models are available at  https://github.com/Tele-AI/TeleStyle. 
}

\maketitle

\vspace{-0.1em}

\section{Introduction}

\begin{figure}
    \centering
     \includegraphics[width=0.9\linewidth]{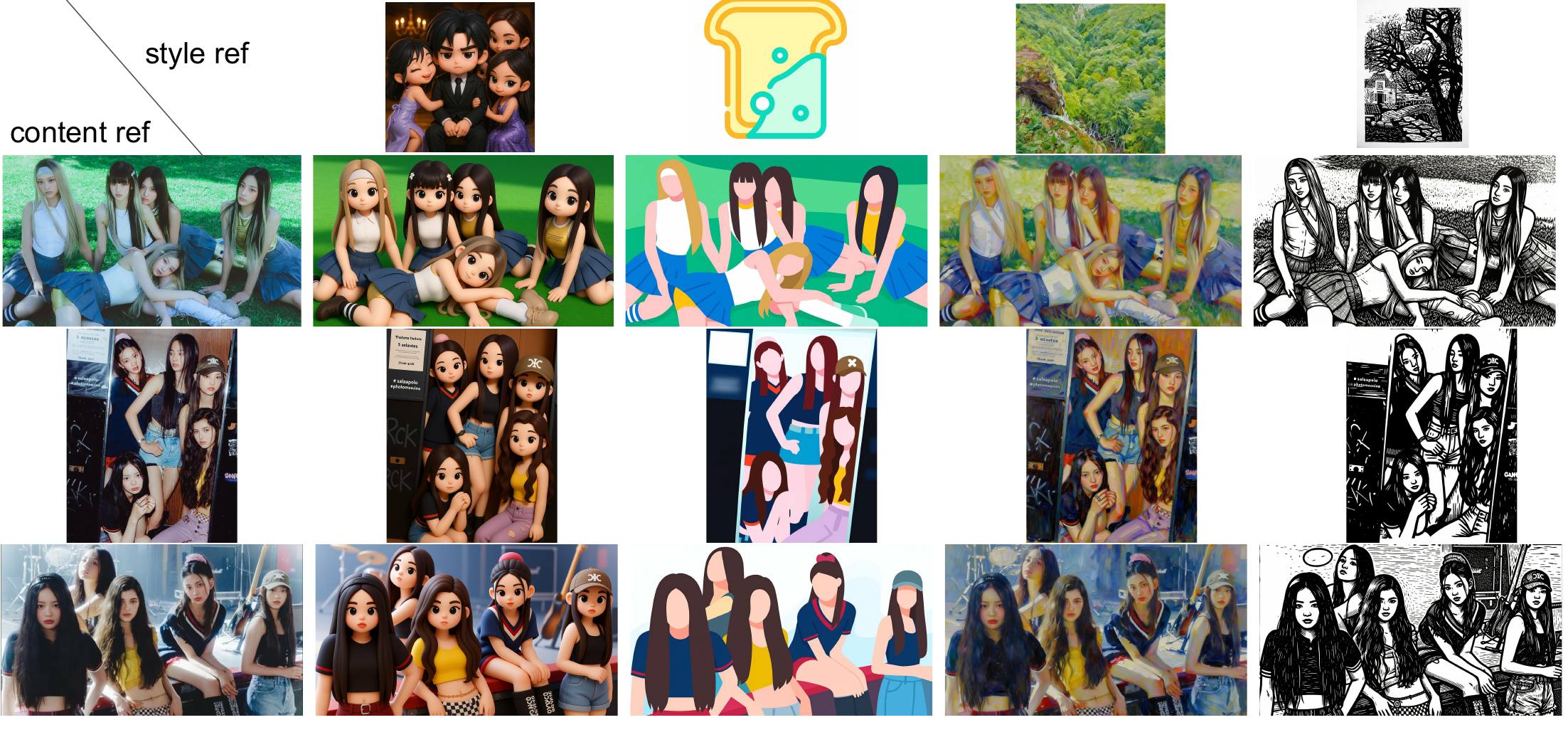}
    \captionsetup{type=figure}
    \captionof{figure}{ TeleStyle-Image accepts style and content references for content-preserving style transfer, while maintaining high aesthetics merit. TeleStyle-Image is the first content-preserving style transfer model built on Qwen-Image-Edit. }
    \label{figure_introduction}
    
\end{figure}

In this technical report, we present TeleStyle, a framework designed to enable content-preserving style transfer within the Qwen-Image-Edit ecosystem \cite{wu2025qwen}. While SDXL-based approaches \cite{podell2023sdxl,wang2023styleadapter} often benefit from the UNet Disentangle Law \cite{zhang2023forgedit} to separate content and style, Diffusion Transformers (DiTs) \cite{dit,sd3} present a unique challenge due to the complex entanglement of features in their internal representations. This makes preserving content structure while transferring style a non-trivial task on DiT architectures.

The Qwen-Image and Qwen-Image-Edit series \cite{wu2025qwen} provide a strong foundation with their state-of-the-art language understanding (Qwen2.5-VL \cite{qwen2.5vl}) and image generation (MMDiT \cite{sd3}). However, they currently lack native support for dual-reference (content and style) stylization. Given that Qwen-Image-Edit employs MS-RoPE \cite{rope} to  distinguish reference inputs, we investigate a data-centric approach to bridge this gap.

We observe that training on limited high-quality triplets restricts generalization, while large-scale "in-the-wild" data introduces noise that degrades content fidelity. To address this trade-off, we propose a Curriculum Continual Learning paradigm \cite{zhang2026qwenstyle}. By progressively training on a hybrid dataset—starting with curated clean triplets to establish fidelity and gradually incorporating synthesized in-the-wild data—we aim to robustly disentangle style from content.

Furthermore, we extend this capability to the temporal domain. We introduce a lightweight video stylization module that conditions on a stylized first frame, propagating the style cues across the sequence to maintain temporal consistency. Figure \ref{figure_introduction} demonstrates TeleStyle's ability to apply diverse styles to both images and videos while respecting their original structural characteristics.

\section{Related Works}
\subsection{Image Style Transfer}

\paragraph{ Zero-Shot Style Transfer with  UNet Diffusion Models}  StyleID \cite{Chung_2024_CVPR} adapters Stable Diffusion \cite{Rombach2021HighResolutionIS} with training-free style injection. StyleShot \cite{gao2025styleshot} designs a style-aware encoder and a content-fusion encoder to conduct test-time-tuning-free style transfer. 
InstantStyle \cite{wang2024instantstyle} uses certain blocks of adapter in the inference process to avoid transferring content of style reference and requires an auxiliary ControlNet \cite{zhang2023adding} to support content reference. CSGO \cite{xing2024csgo} projects content reference into UNet Encoder and style reference into UNet Decoder with content-style image pairs for training.  

\paragraph{Zero-Shot Style Transfer with Conditional DiT}  SD3 \cite{esser2024scaling} and FLUX \cite{flux2024} improves text-to-image task significantly by scaling up DiT \cite{dit} parameters. With these new powerful text-to-image DiT models, OminiControl \cite{tan2024ominicontrol} and EasyControl \cite{zhang2025easycontrol} enable Conditional Image Generation by concatenating condition image with text condition and noisy latent in self attention modules.   OmniStyle \cite{wang2025omnistyle} constructs and filters [style ref, content ref, target] triplets to train Conditional DiT with style transfer capability. DreamO \cite{mou2025dreamo} unifies multiple conditional image generation tasks into one DiT model, including content preservation and style transfer. OmniGen-v2 \cite{wu2025omnigen2} enables in-context image generation by leveraging hidden states of    interleaved images and texts tokens generated by an MLLM  as input to the diffusion decoder.  Qwen Image Edit \cite{wu2025qwen} could handle multiple reference images for subject-driven customization. However, currently Qwen Image Edit does not support subject+style references, neither does FLUX-Kontext \cite{labs2025flux}. OmniConsistency \cite{song2025omniconsistency} trained a separate content consistency branch and relies on external Style Loras \cite{lora} to conduct style transfer with content preservation. Instead, our model unifies content preservation and style transfer capability in one unified model, which is capable to handle universal style categories without the need for Style Loras trained on one specific style.
\subsection{Video Stylization}

Video stylization has advanced rapidly with the rise of generative models. Existing methods fall into three broad categories: training-free plug-and-play approaches, end-to-end models trained on paired data, and unified frameworks supporting multiple conditioning modalities.

Training-free methods typically propagate stylization from a key frame across the entire video. For instance, AnyV2V~\cite{ku2024anyv2v} generates edited videos by injecting features from a user-provided stylized first frame, though it relies on an external image stylization model. TokenFlow~\cite{qu2025tokenflow} proposes a dual-codebook tokenizer that unifies high-level semantic understanding and fine-grained visual generation through shared discrete tokens—though it is not specifically designed for video stylization.These approaches inherit style representations from pre-trained generative priors, where style is entangled with latent semantics rather than aligned via paired supervision. Consequently, the extracted style features may exhibit misalignment between semantic structure and low-level texture details, leading to artifacts or inconsistent stylization.

End-to-end approaches leverage paired style data for supervised training. StyleCrafter~\cite{liu2023stylecrafter} injects CLIP-derived style features into a U-Net denoiser, while StyleMaster~\cite{ye2025stylemaster} enhances DiT backbones with dedicated global and local style extractors to better preserve texture fidelity and prevent content leakage. PickStyle~\cite{mehraban2025pickstyle} inserts low-rank style adapters into pre-trained video diffusion models and trains on synthetic clips derived from paired still images. DreamStyle~\cite{li2026dreamstyle} presents a unified framework supporting text-, style-image-, and first-frame-guided stylization, using token-specific LoRA to disentangle heterogeneous conditioning signals and achieve strong cross-task generalization.

Recent large-scale efforts further enable general-purpose video editing with stylization capabilities. Señorita-2M~\cite{zi2025senorita} introduces a dataset of 2 million high-quality video editing pairs, mitigating data scarcity in supervised learning. Ditto~\cite{bai2025ditto} scales instruction-based editing via Ditto-1M—a synthetic dataset containing abundant style transfer examples generated using depth-aware video synthesis. LucyEdit~\cite{decart2025lucyedit} adopts a dual-branch diffusion architecture with shared temporal layers to jointly preserve subject identity and apply mask-free, text-guided edits.
Despite their versatility, these unified models often underperform specialized stylization methods in terms of style fidelity and fine-grained texture control, highlighting a trade-off between generality and expert-level stylization quality.

\begin{figure}
    \centering
    \includegraphics[
        trim=0cm 17cm 7cm 0cm,
        clip,
        width=1.0\linewidth
    ]{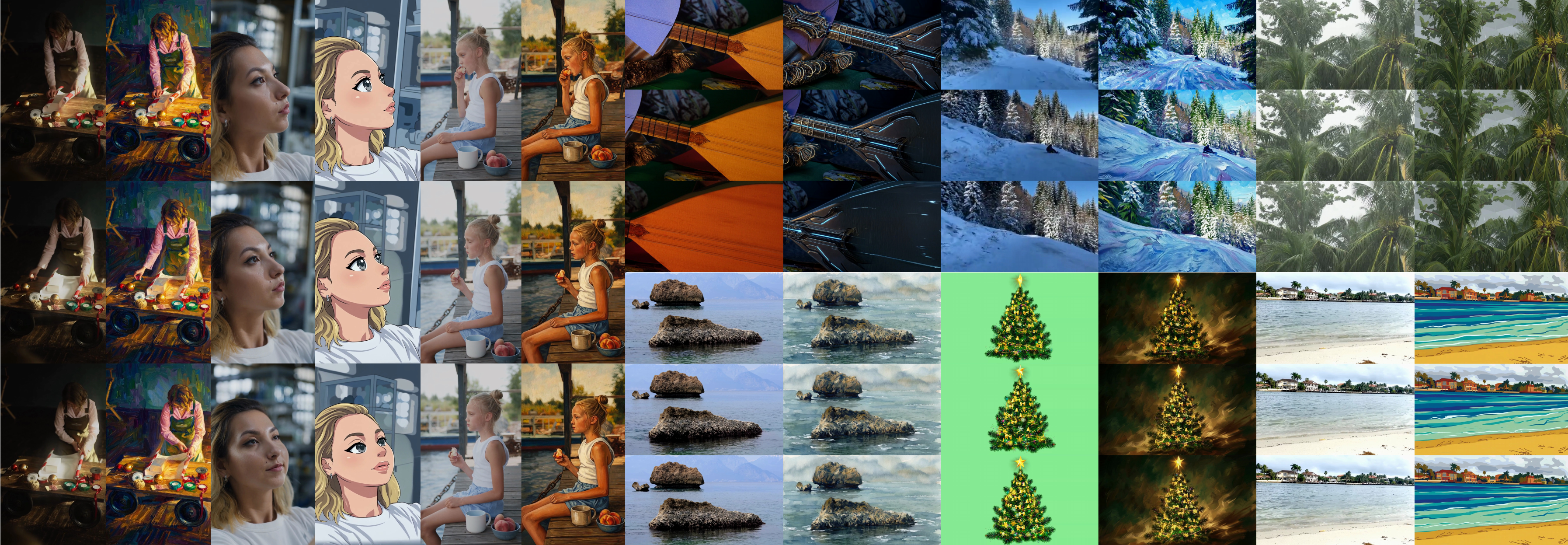}
    \captionsetup{type=figure}
    \captionof{figure}{ TeleStyle-Video takes a stylized first frame and a source video as input and propagates the style coherently across the entire sequence. It achieves high-quality results across a diverse range of artistic styles—including oil painting, sci-fi, and watercolor—while preserving temporal consistency. Notably, it remains effective even under challenging conditions such as anime-style rendering, where large structural and contour variations are present. }
    \label{figure_introduction}
    
\end{figure}

\section{Methods}

\subsection{Content-Preserving Style Transfer on Images}

\subsubsection{Triplet Training Dataset Construction}

\begin{figure*}[t]
  \centering
  \includegraphics[width=0.8\linewidth]{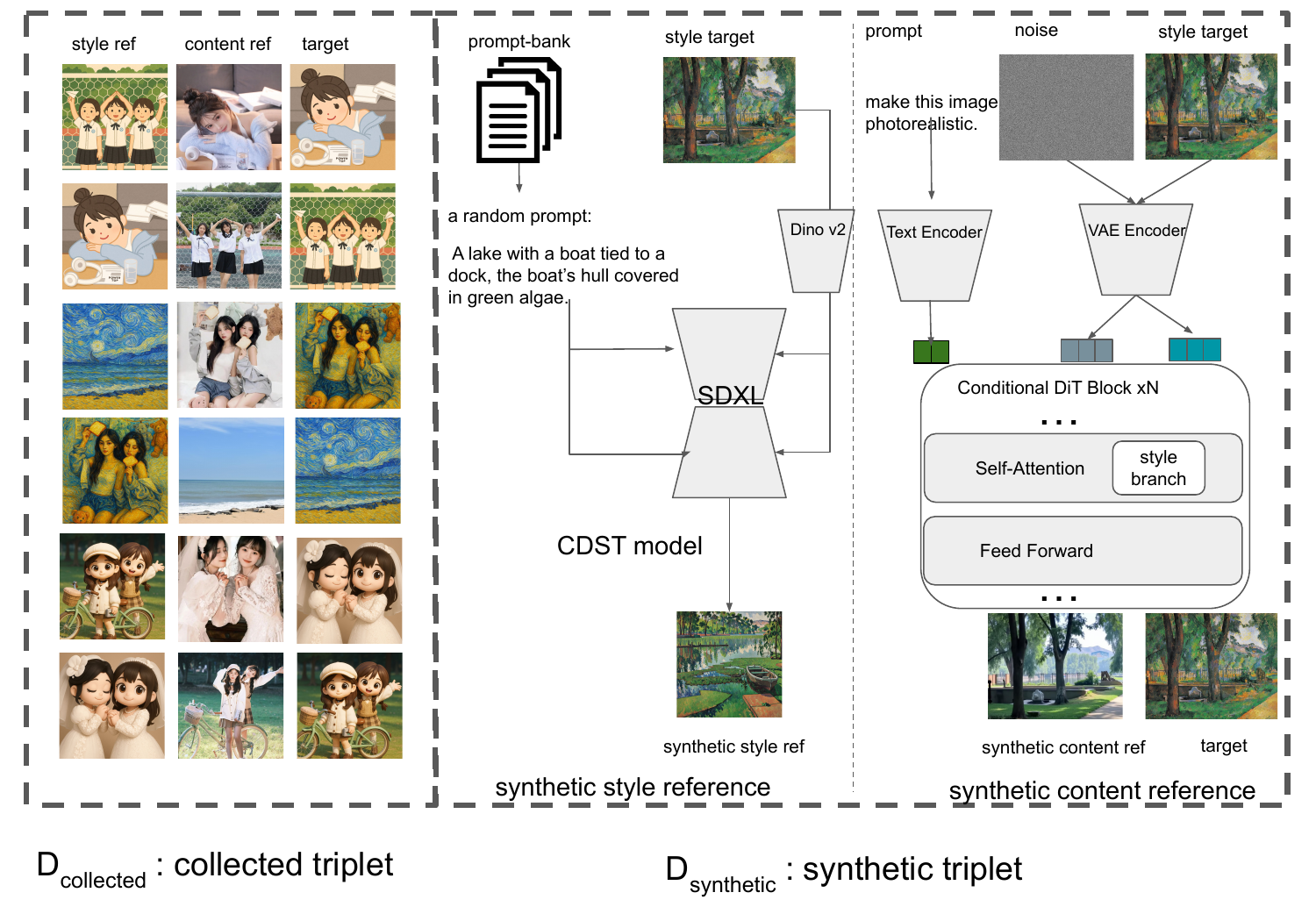}
  \caption{Overview of dataset construction: Collected triplets $D_{collected}$ (left) and synthetic triplets $D_{synthetic}$ (right).}
  \vspace{-0.2cm}
  \label{figure_triplet}
\end{figure*}

Unlike subject-driven image data \cite{tan2024ominicontrol,zhang2025easycontrol}, which naturally occurs in video frames or photo albums, explicit style transfer triplets (Style, Content, Target) are scarce in real-world scenarios. To address this, we constructed a hybrid dataset comprising both collected and synthesized triplets.

First, we constructed a clean dataset, $D_{collected}$, by sampling triplets from \cite{song2025omniconsistency}, generated via GPT-4o \cite{gpt4o}, and supplementing them with LoRA-generated samples from the open-source community. These samples underwent rigorous data filtering. However, this process is resource-intensive, yielding only 30 distinct style categories. Models trained solely on this limited distribution generalized poorly to unseen styles.

To overcome this bottleneck, we introduced a reverse triplet synthesis framework inspired by \cite{wang2023stylediffusion} to generate training data from in-the-wild style images \cite{li2024styletokenizer}. As shown in Figure \ref{figure_triplet}, our synthesis pipeline operates as follows:
\begin{enumerate}
    \item \textbf{Content Reference Generation:} Given a stylized target image, we convert it into a photorealistic content reference using a pre-trained image editing model based on FLUX.1-dev \cite{flux2024}.
    \item \textbf{Style Reference Generation:} We generate a corresponding style reference using CDST \cite{zhang2025cdst}, which leverages DINOv2 \cite{oquab2023dinov2} for style extraction.
    \item \textbf{Prompt Strategy:} We sample random prompts from a pre-defined bank. Notably, we exclude humans from these prompts, as we observed that SDXL-based CDST suffers from identity leakage, which degrades the separation between content and style.
\end{enumerate}

By matching stylized images within the same style clusters (assigning one as the style reference and the other as the target), we generated 1 million synthetic triplets, denoted as $D_{synthetic}$. This complements the 300k high-quality triplets in $D_{collected}$.

\subsubsection{Content-Preserving Style Transfer via Curriculum Continual Learning}

Given the disparity in data quality and the varying difficulty of style categories, naive training on the mixed dataset proved suboptimal. To address this, we propose a three-stage \textit{Curriculum Continual Learning} framework for TeleStyle.

\paragraph{Stage 1: Capability Activation ($D_1$).}
We first train the model, denoted as $Q_1$, on the clean dataset $D_{collected}$ (referred to here as $D_1$). This stage activates the fundamental content-preserving style transfer capability of Qwen-Image-Edit. However, $Q_1$ exhibits two primary limitations:
\begin{itemize}
    \item \textbf{Detail Loss:} It fails to preserve subtle content characteristics, such as specific facial identities in multi-person scenes.
    \item \textbf{Poor Generalization:} It struggles to adapt to out-of-distribution (OOD) styles not present in $D_{collected}$.
\end{itemize}

\paragraph{Stage 2: Content Fidelity Refinement ($D_2$).}
To address detail loss, we curate a subset of $D_{collected}$ by filtering for high content consistency, increasing the sampling weight of these high-fidelity triplets. Initialized with $Q_1$, we fine-tune the model on this refined dataset ($D_2$) to obtain $Q_2$. Validation results demonstrate that $Q_2$ preserves content characteristics with significantly higher precision than $Q_1$, effectively resolving the detail loss issue.

\paragraph{Stage 3: Robust Generalization ($D_3$).}
Finally, we aim to improve style generalization without compromising the content fidelity achieved in Stage 2. While $D_{synthetic}$ provides vast stylistic diversity, its content references often contain structural noise (e.g., altered small objects or facial features). Training on $D_{synthetic}$ alone leads to content degradation.
To mitigate catastrophic forgetting, we construct $D_3$ by mixing $D_2$ with a carefully tuned, low ratio of $D_{synthetic}$. Initializing with $Q_2$, we train the final model $Q_3$. This strategy significantly enhances generalization to OOD styles while maintaining acceptable content consistency.

\begin{algorithm}[t]
  \caption{TeleStyle Training: Curriculum Continual Learning} \label{alg:Style-CCL}
  \small
  \begin{algorithmic}[1]
    \vspace{.04in}
    \State {\bf Input}:
    \State \quad $D_1$: Triplets from natural distribution $D_{collected}$.
    \State \quad $D_2$: $D_{collected}$ re-weighted for high content consistency.
    \State \quad $D_3$: $D_2$ mixed with a low ratio of $D_{synthetic}$.
    \State {\bf Output}: TeleStyle-Image Model ($Q_3$)
    \State
    \State {\bf Procedure}:
    \State \quad Train LoRA $Q_1 \leftarrow \{D_1, \text{Qwen-Image-Edit}\}$
    \State \quad Train LoRA $Q_2 \leftarrow \{D_2, Q_1\}$
    \State \quad Train LoRA $Q_3 \leftarrow \{D_3, Q_2\}$
    \State \textbf{return} $Q_3$
  \end{algorithmic}
\end{algorithm}

\subsubsection{Training and Inference Details}

We employ a Low-Rank Adaptation (LoRA) \cite{lora} approach for training, as fine-tuning all parameters of the Qwen-Image MMDiT \cite{wu2025qwen} yielded negligible performance gains. The optimization objective is based on rectified flow-matching \cite{rectifiedflow,flowmatching}:
\begin{equation}
L = \mathbb{E}_{t, \epsilon \sim \mathcal{N}(0, I)} \left\| v_\theta(x_t, t, c_{style}, c_{content}, c_{prompt}) - (\epsilon - x_0) \right\|_2^2
\end{equation}
where $x_t$ represents the image features at time $t$; $c_{style}$, $c_{content}$, and $c_{prompt}$ are the conditioning inputs; $v_\theta$ denotes the velocity field; $x_0$ is the target image feature; and $\epsilon$ is the noise.

\textbf{Prompting Strategy:} During training, we utilize a standardized prompt template:
\begin{quote}
    \textit{"Style Transfer the style of Figure 2 to Figure 1, and keep the content and characteristics of Figure 1."}
\end{quote}
While TeleStyle supports explicit style descriptors during inference (e.g., \textit{"Transfer Figure 1 into Van Gogh style"}), we recommend using the default template for optimal stability. All results presented in this report were generated using the default prompt.

\textbf{Inference Configuration:} We observed that maintaining identical aspect ratios between the content reference and the generated output is crucial. The style reference is resized to a square of dimension $\min(H_{target}, W_{target})$.

\subsection{Video Stylization}

To extend our framework to the temporal domain, we developed a first-frame-conditioned propagation model. This model transfers the visual style of a reference image to an entire video sequence while preserving the original motion dynamics.

Formally, given a source content video $V = \{ \mathbf{v}_0, \mathbf{v}_1, \dots, \mathbf{v}_{T-1} \}$ and a style reference $I = \mathbf{v}_0^{\text{style}}$, our model $M_\theta$ generates a stylized video $\hat{V} = \{ \hat{\mathbf{v}}_0, \hat{\mathbf{v}}_1, \dots, \hat{\mathbf{v}}_{T-1} \}$ such that:
\[
\hat{\mathbf{v}}_0 \approx I, \quad \text{and} \quad \text{motion}(\hat{V}) \approx \text{motion}(V)
\]
where $\text{motion}(\cdot)$ denotes the temporal trajectory in pixel or latent space.

\begin{figure*}[t]
  \centering
  \includegraphics[width=0.8\linewidth]{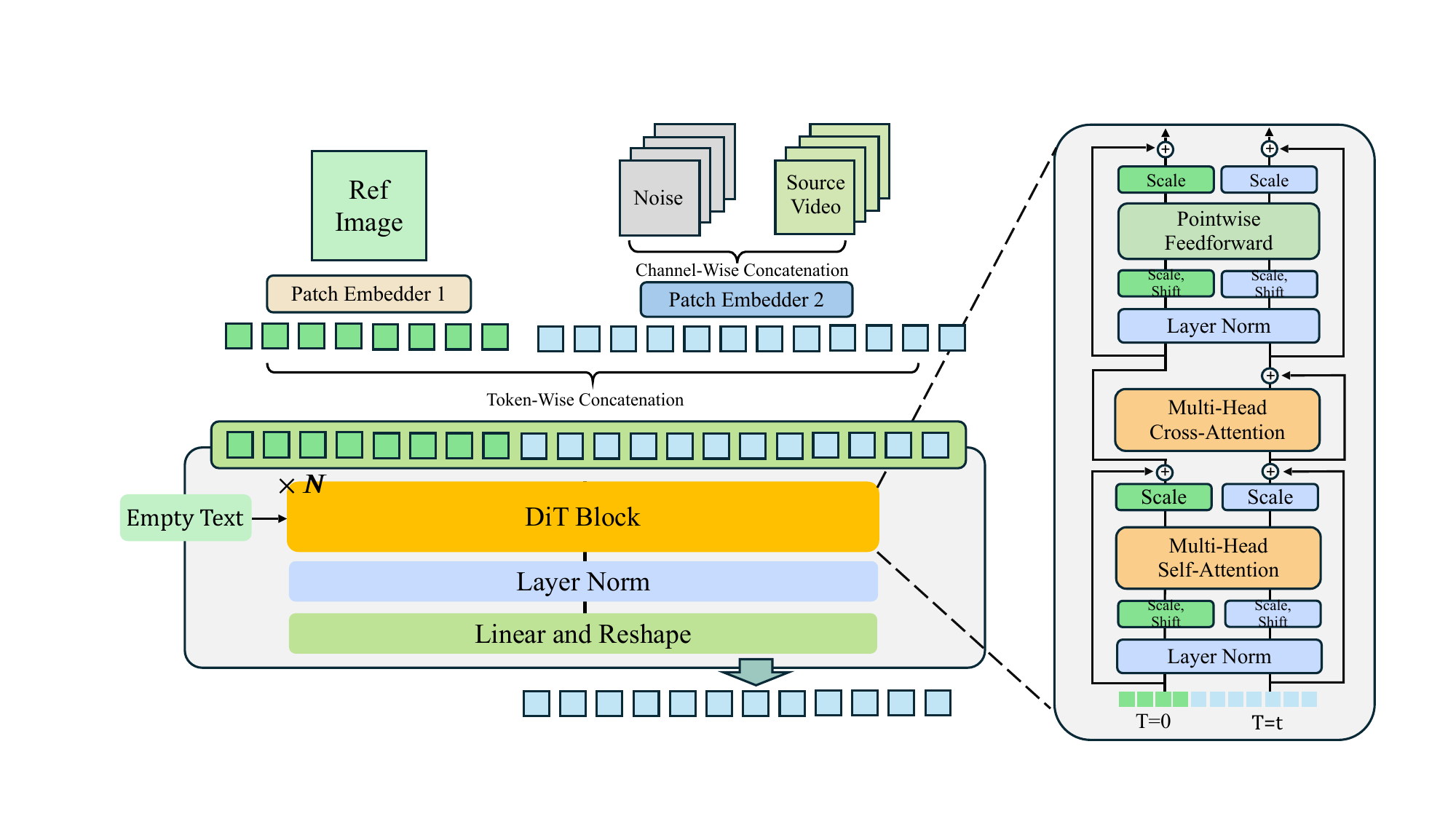}
  \caption{\textbf{Overview of the video stylization architecture.} A reference image and source video frames are encoded via dual Patch Embedders, fused with noisy latents, and processed by DiT blocks alongside empty text tokens.}
  \vspace{-0.2cm}
  \label{figure_structure}
\end{figure*}

\textbf{Architecture:} Our model is built upon the Wan2.1-1.3B \cite{wan2025} backbone, integrating architectural insights from FullDiT \cite{ju2025fulldit}. As illustrated in Figure \ref{figure_structure}, we employ a multimodal conditional Diffusion Transformer (DiT). To streamline the pipeline, we bypass textual conditioning by using empty text prompts; all necessary semantic signals (style, structure, motion) are encoded directly from the visual inputs.

\textbf{Input Processing:} The inputs include:
\begin{itemize}
    \item A style reference image $I \in \mathbb{R}^{H \times W \times C}$.
    \item A sequence of noisy video frames $\tilde{V}_t = \{ \tilde{\mathbf{v}}_t^{(i)} \}_{i=0}^{T-1}$ at timestep $t$.
    \item An empty text embedding $\mathbf{e}_{\emptyset}$.
\end{itemize}
Two dedicated Patch Embedders encode the style image and video frames into spatial token sequences $\mathbf{Z}_I$ and $\mathbf{Z}_V$, respectively. These are concatenated channel-wise with the noisy latent representation $\tilde{\mathbf{z}}_t$, fused with the empty text embedding, and processed by $N$ DiT blocks.

\textbf{Temporal Alignment:} Crucially, we assign the style reference a temporal index of 0, while the noisy video tokens retain their original frame indices $\{0, 1, \dots, T-1\}$. This positional encoding strategy enables the model to treat the style reference as a temporal anchor, facilitating consistent style propagation.

\textbf{Optimization:} We train the model using a flow matching objective:
\[
\mathcal{L}_{\text{FM}} = \mathbb{E}_{t,\mathbf{x}_0,\mathbf{x}_1}\left[ \left\| \mathbf{v}_\theta(\mathbf{x}_t, t) - \mathbf{u}_t(\mathbf{x}_t) \right\|^2 \right]
\]
where $\mathbf{x}_t = (1 - t)\mathbf{x}_0 + t\mathbf{x}_1$ is the linear interpolation between the ground-truth stylized video $\mathbf{x}_0$ and noise $\mathbf{x}_1 \sim \mathcal{N}(0, \mathbf{I})$. The vector field $\mathbf{u}_t(\mathbf{x}_t) = \mathbf{x}_1 - \mathbf{x}_0$ guides the velocity predictor $\mathbf{v}_\theta$. Trained on large-scale synthesized video stylization pairs, this approach yields temporally coherent results without requiring optical flow guidance or test-time optimization.

\section{Experiments and Discussion}

\subsection{Implementation and Evaluation}
\paragraph{Implementation Details.} TeleStyle-Image adopt Lora \cite{lora} to Qwen-Image-Edit-2509 \cite{wu2025qwen} to train TeleStyle V1. The ranks for Lora is 32.  Gradient Checkpointing \cite{griewank2000algorithm} is applied to save memory and the model is trained with min-edge=1024 . Our model is trained with 4 H100 GPUs, batch size is 1 for each GPU, learning rate is 1e-4.

TeleStyle-Video is trained on a combination of the Ditto dataset~\cite{bai2025ditto} and an internal video collection. However, since Ditto videos are synthesized under depth-consistent conditions, many clips exhibit significant texture changes without meaningful motion, which can lead to temporal misalignment during stylization. To address this, we filter the Ditto data using CLIP-based frame similarity~\cite{radford2021learningclipsim}, discarding clips with negligible motion to ensure temporally coherent supervision. The model is trained on 8 H100 GPUs with a batch size of 4 for each GPU and a learning rate of $1 \times 10^{-5}$.

\paragraph{Evaluation Benchmark.}
We select 50 style references and 40 content references of different ratios, mutually pair each of them to generate 2000 style-content pairs for testing.  We further select 10 style references and 10 content references as  validation set.  The style references cover diverse  style genres and the content references include different number of persons with diverse gestures, scenes/buildings and subjects in complex scenarios.
\paragraph{Evaluation Metrics.} 
We evaluate our method with the following metrics. 
For \textbf{Style Consistency}, we use  CSD Score~\cite{csd} to measure the style similarity between the style reference and the generated image. 
For \textbf{Aesthetics},we use the LAION Aesthetics Predictor~\cite{schuhmann2022laion} to estimate the aesthetic quality of the generated image. For \textbf{Content Preservation},  we propose a new \emph{Content Preservation Cut-Off Score} (CPC Score), which augments the original Content Preservation Score~\cite{wang2025omnistyle} with a style consistency threshold. Intuitively, a model that simply replicates the content reference without transferring style would receive an artificially high content score. To avoid this, we first use Qwen-VL~\cite{qwen2.5vl} to generate a detailed caption $T_\text{vlm}$ for the content reference image $I_{\text{content}}$, and compute the CLIP score~\cite{Radford2021LearningTV} between $T_{\text{vlm}}$ and the generated image $I_{\text{res}}$. We then compute the CSD Score between $I_{\text{style}}$ and $I_{\text{res}}$; if this score falls below a threshold, the CLIP score is set to zero as a penalty. 

\begin{equation}
\resizebox{0.6\columnwidth}{!}{
$CPC@thresh =
\begin{cases}
CLIP(I_{res},T_{vlm}), & \text{if }  CSD(I_{res},I_{style})>=\text{thresh} \\
0, & \text{if } CSD(I_{res},I_{style})<\text{thresh} \\
\end{cases}$

}
\end{equation}

\subsection{Comparison with State-of-the-art Methods}
\begin{figure*}[!htb]
  \centering
   \includegraphics[width=0.9\linewidth]{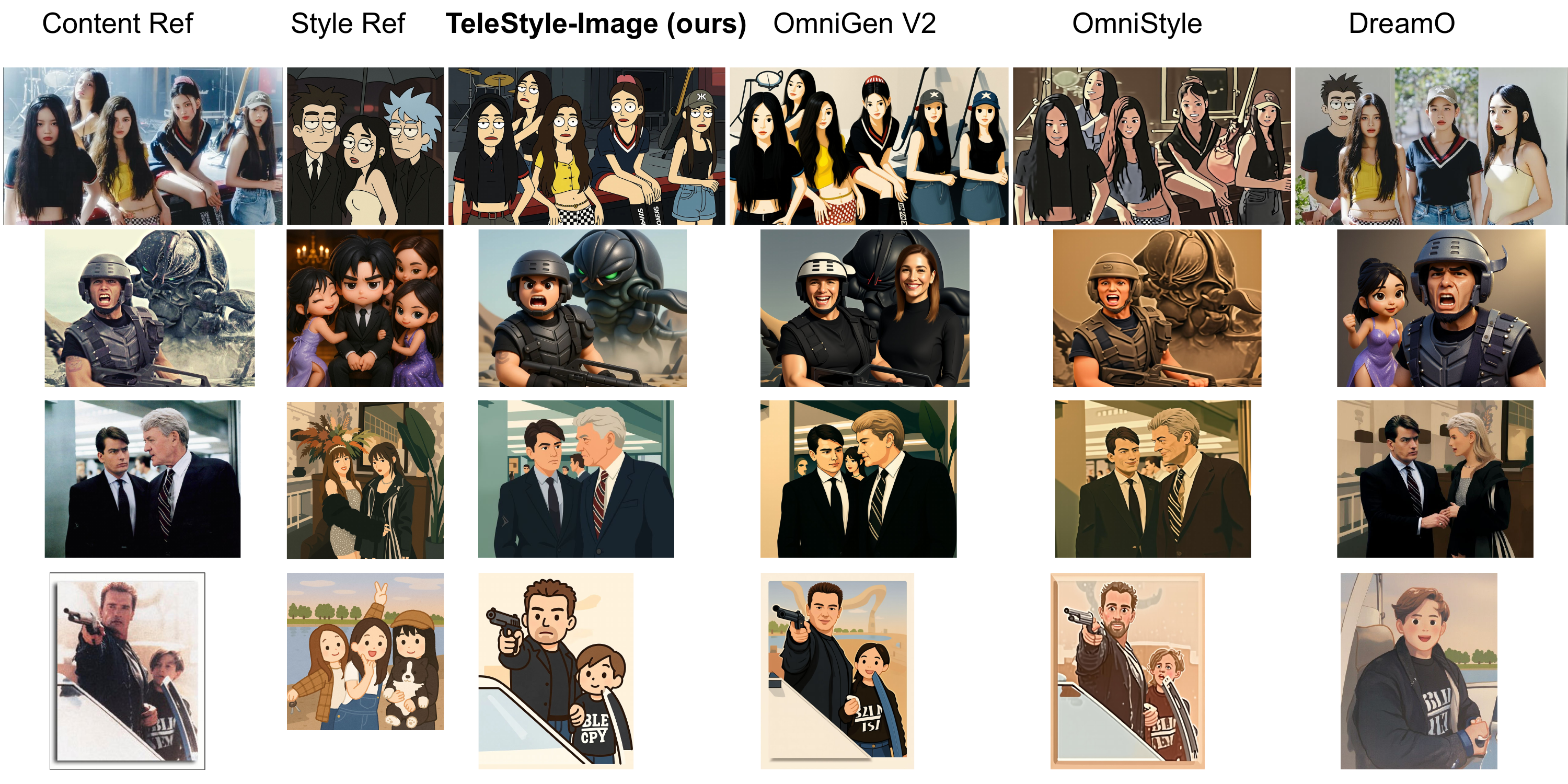}
   \caption{ Qualitative Comparison with State-of-the-art Style Transfer Models.   }
   \label{figure_sota}
\end{figure*}
\begin{table*}[t]
\scriptsize
\begin{tabular}{c|c|c|c|c}
\hline
{} &  {Style Similarity }  & {Content Preservation}  & {Content Preservation }  & { }  \\
{Methods} &  {CSD Score$\uparrow$}  & { CPC Score@0.5 $\uparrow$}  & {CPC Score@0.3:0.9 $\uparrow$}  & {Aesthetic Score$\uparrow$}  \\
\hline
StyleID \cite{Chung_2024_CVPR} & 0.453 & 0.190 & 0.180& 5.749 \\
StyleShot \cite{gao2025styleshot} & 0.450 & 0.227 & 0.116& 5.740 \\
InstantStyle \cite{wang2024instantstyle} & 0.397 & 0.189 &0.134& 5.464 \\
CSGO \cite{xing2024csgo} & \underline{0.535} & \underline{0.379} & \underline{0.224} & 5.969 \\
OmniStyle \cite{wang2025omnistyle}& {0.447} & 0.194 & 0.163 & 5.881  \\
OmniGen-v2 \cite{wu2025omnigen2} & 0.462 & 0.243 & 0.166& 5.843\\
DreamO \cite{mou2025dreamo} & 0.402 & 0.193 & 0.102 & \underline{6.149} \\

\hline

{\bf TeleStyle (ours)} & {\bf 0.577} & {\bf 0.441} & {\bf 0.304}&  {\bf 6.317} \\

\hline
\end{tabular}
\caption{Quantitative comparison of our TeleStyle with previous state-of-the-art style transfer methods. }
\label{table_sota}
\end{table*}

\paragraph{Quantitative Comparison}
We quantitatively compare our TeleStyle V1  with multiple DiT-based style transfer models in Table \ref{table_sota}, from the aspects of style similarity, content preservation and aesthetics score. Our TeleStyle V1 demonstrates significant advantages over previous models.

\paragraph{Qualitative Comparison}
We present qualitative visual comparison with some representative DiT-based style transfer models  in Figure \ref{figure_sota}, where our TeleStyle V1 demonstrates superior style similarity and content consistency than previous models, while maintaining high aesthetics values. 



\section{Conclusion}
In this work, we presented TeleStyle, a practical framework for content-preserving style transfer across both images and videos. By addressing the feature entanglement challenges inherent in Diffusion Transformers (DiTs), we successfully unlocked the potential of the Qwen-Image-Edit and Wan2.1 architectures for precise stylization tasks. Experimental results demonstrate that TeleStyle achieves state-of-the-art performance in style similarity, content preservation, and aesthetic quality. We hope this work serves as a robust baseline for future research in DiT-based style transfer and high-fidelity video editing.

\bibliography{paper}

\begin{thebibliography}{}

\bibitem[\protect\citename{Bai {\em et~al.}, }2025a]{bai2025ditto}
Bai, Qingyan, Wang, Qiuyu, Ouyang, Hao, Yu, Yue, Wang, Hanlin, Wang, Wen, Cheng, Ka~Leong, Ma, Shuailei, Zeng, Yanhong, Liu, Zichen, {\em et~al.} 2025a.
\newblock Scaling instruction-based video editing with a high-quality synthetic dataset.
\newblock {\em arXiv preprint arXiv:2510.15742}.

\bibitem[\protect\citename{Bai {\em et~al.}, }2025b]{qwen2.5vl}
Bai, Shuai, Chen, Keqin, Liu, Xuejing, Wang, Jialin, Ge, Wenbin, Song, Sibo, Dang, Kai, Wang, Peng, Wang, Shijie, Tang, Jun, {\em et~al.} 2025b.
\newblock Qwen2. 5-vl technical report.
\newblock {\em arXiv preprint arXiv:2502.13923}.

\bibitem[\protect\citename{Chung {\em et~al.}, }2024]{Chung_2024_CVPR}
Chung, Jiwoo, Hyun, Sangeek, \& Heo, Jae-Pil. 2024 (June).
\newblock Style Injection in Diffusion: A Training-free Approach for Adapting Large-scale Diffusion Models for Style Transfer.
\newblock {\em Pages  8795--8805 of:} {\em Proceedings of the IEEE/CVF Conference on Computer Vision and Pattern Recognition (CVPR)}.

\bibitem[\protect\citename{Esser {\em et~al.}, }2024a]{sd3}
Esser, Patrick, Kulal, Sumith, Blattmann, Andreas, Entezari, Rahim, M{\"u}ller, Jonas, Saini, Harry, Levi, Yam, Lorenz, Dominik, Sauer, Axel, Boesel, Frederic, {\em et~al.} 2024a.
\newblock Scaling rectified flow transformers for high-resolution image synthesis.
\newblock {\em In:} {\em Forty-first international conference on machine learning}.

\bibitem[\protect\citename{Esser {\em et~al.}, }2024b]{esser2024scaling}
Esser, Patrick, Kulal, Sumith, Blattmann, Andreas, Entezari, Rahim, M{\"u}ller, Jonas, Saini, Harry, Levi, Yam, Lorenz, Dominik, Sauer, Axel, Boesel, Frederic, {\em et~al.} 2024b.
\newblock Scaling rectified flow transformers for high-resolution image synthesis.
\newblock {\em In:} {\em Forty-first international conference on machine learning}.

\bibitem[\protect\citename{Gao {\em et~al.}, }2025]{gao2025styleshot}
Gao, Junyao, Sun, Yanan, Liu, Yanchen, Tang, Yinhao, Zeng, Yanhong, Qi, Ding, Chen, Kai, \& Zhao, Cairong. 2025.
\newblock Styleshot: A snapshot on any style.
\newblock {\em IEEE Transactions on Pattern Analysis and Machine Intelligence}.

\bibitem[\protect\citename{Griewank \& Walther, }2000]{griewank2000algorithm}
Griewank, Andreas, \& Walther, Andrea. 2000.
\newblock Algorithm 799: revolve: an implementation of checkpointing for the reverse or adjoint mode of computational differentiation.
\newblock {\em ACM Transactions on Mathematical Software (TOMS)}, {\bf 26}(1), 19--45.

\bibitem[\protect\citename{Hu {\em et~al.}, }2021]{lora}
Hu, Edward~J, Shen, Yelong, Wallis, Phillip, Allen-Zhu, Zeyuan, Li, Yuanzhi, Wang, Shean, Wang, Lu, \& Chen, Weizhu. 2021.
\newblock Lora: Low-rank adaptation of large language models.
\newblock {\em arXiv preprint arXiv:2106.09685}.

\bibitem[\protect\citename{Hurst {\em et~al.}, }2024]{gpt4o}
Hurst, Aaron, Lerer, Adam, Goucher, Adam~P, Perelman, Adam, Ramesh, Aditya, Clark, Aidan, Ostrow, AJ, Welihinda, Akila, Hayes, Alan, Radford, Alec, {\em et~al.} 2024.
\newblock Gpt-4o system card.
\newblock {\em arXiv preprint arXiv:2410.21276}.

\bibitem[\protect\citename{Ju {\em et~al.}, }2025]{ju2025fulldit}
Ju, Xuan, Ye, Weicai, Liu, Quande, Wang, Qiulin, Wang, Xintao, Wan, Pengfei, Zhang, Di, Gai, Kun, \& Xu, Qiang. 2025.
\newblock Fulldit: Multi-task video generative foundation model with full attention.
\newblock {\em arXiv preprint arXiv:2503.19907}.

\bibitem[\protect\citename{Ku {\em et~al.}, }2024]{ku2024anyv2v}
Ku, Max, Wei, Cong, Ren, Weiming, Yang, Harry, \& Chen, Wenhu. 2024.
\newblock Anyv2v: A tuning-free framework for any video-to-video editing tasks.
\newblock {\em arXiv preprint arXiv:2403.14468}.

\bibitem[\protect\citename{Labs, }2024]{flux2024}
Labs, Black~Forest. 2024.
\newblock {\em FLUX}.
\newblock \url{https://github.com/black-forest-labs/flux}.

\bibitem[\protect\citename{Labs {\em et~al.}, }2025]{labs2025flux}
Labs, Black~Forest, Batifol, Stephen, Blattmann, Andreas, Boesel, Frederic, Consul, Saksham, Diagne, Cyril, Dockhorn, Tim, English, Jack, English, Zion, Esser, Patrick, {\em et~al.} 2025.
\newblock FLUX. 1 Kontext: Flow Matching for In-Context Image Generation and Editing in Latent Space.
\newblock {\em arXiv preprint arXiv:2506.15742}.

\bibitem[\protect\citename{Li {\em et~al.}, }2026]{li2026dreamstyle}
Li, Mengtian, Chen, Jinshu, Zhao, Songtao, Feng, Wanquan, Tu, Pengqi, \& He, Qian. 2026.
\newblock DreamStyle: A Unified Framework for Video Stylization.
\newblock {\em arXiv preprint arXiv:2601.02785}.

\bibitem[\protect\citename{Li {\em et~al.}, }2024]{li2024styletokenizer}
Li, Wen, Fang, Muyuan, Zou, Cheng, Gong, Biao, Zheng, Ruobing, Wang, Meng, Chen, Jingdong, \& Yang, Ming. 2024.
\newblock Styletokenizer: Defining image style by a single instance for controlling diffusion models.
\newblock {\em Pages  110--126 of:} {\em European Conference on Computer Vision}.
\newblock Springer.

\bibitem[\protect\citename{Lipman {\em et~al.}, }2022]{flowmatching}
Lipman, Yaron, Chen, Ricky~TQ, Ben-Hamu, Heli, Nickel, Maximilian, \& Le, Matt. 2022.
\newblock Flow matching for generative modeling.
\newblock {\em arXiv preprint arXiv:2210.02747}.

\bibitem[\protect\citename{Liu {\em et~al.}, }2023]{liu2023stylecrafter}
Liu, Gongye, Xia, Menghan, Zhang, Yong, Chen, Haoxin, Xing, Jinbo, Wang, Yibo, Wang, Xintao, Yang, Yujiu, \& Shan, Ying. 2023.
\newblock Stylecrafter: Enhancing stylized text-to-video generation with style adapter.
\newblock {\em arXiv preprint arXiv:2312.00330}.

\bibitem[\protect\citename{Liu {\em et~al.}, }2022]{rectifiedflow}
Liu, Xingchao, Gong, Chengyue, \& Liu, Qiang. 2022.
\newblock Flow straight and fast: Learning to generate and transfer data with rectified flow.
\newblock {\em arXiv preprint arXiv:2209.03003}.

\bibitem[\protect\citename{Mehraban {\em et~al.}, }2025]{mehraban2025pickstyle}
Mehraban, Soroush, Adeli, Vida, Rommann, Jacob, Taati, Babak, \& Truskovskyi, Kyryl. 2025.
\newblock PickStyle: Video-to-Video Style Transfer with Context-Style Adapters.
\newblock {\em arXiv preprint arXiv:2510.07546}.

\bibitem[\protect\citename{Mou {\em et~al.}, }2025]{mou2025dreamo}
Mou, Chong, Wu, Yanze, Wu, Wenxu, Guo, Zinan, Zhang, Pengze, Cheng, Yufeng, Luo, Yiming, Ding, Fei, Zhang, Shiwen, Li, Xinghui, {\em et~al.} 2025.
\newblock Dreamo: A unified framework for image customization.
\newblock {\em arXiv preprint arXiv:2504.16915}.

\bibitem[\protect\citename{Oquab {\em et~al.}, }2023]{oquab2023dinov2}
Oquab, Maxime, Darcet, Timoth{\'e}e, Moutakanni, Th{\'e}o, Vo, Huy, Szafraniec, Marc, Khalidov, Vasil, Fernandez, Pierre, Haziza, Daniel, Massa, Francisco, El-Nouby, Alaaeldin, {\em et~al.} 2023.
\newblock Dinov2: Learning robust visual features without supervision.
\newblock {\em arXiv preprint arXiv:2304.07193}.

\bibitem[\protect\citename{Peebles \& Xie, }2023]{dit}
Peebles, William, \& Xie, Saining. 2023.
\newblock Scalable Diffusion Models with Transformers.
\newblock {\em In:} {\em ICCV}.

\bibitem[\protect\citename{Podell {\em et~al.}, }2023]{podell2023sdxl}
Podell, Dustin, English, Zion, Lacey, Kyle, Blattmann, Andreas, Dockhorn, Tim, M{\"u}ller, Jonas, Penna, Joe, \& Rombach, Robin. 2023.
\newblock Sdxl: Improving latent diffusion models for high-resolution image synthesis.
\newblock {\em arXiv preprint arXiv:2307.01952}.

\bibitem[\protect\citename{Qu {\em et~al.}, }2025]{qu2025tokenflow}
Qu, Liao, Zhang, Huichao, Liu, Yiheng, Wang, Xu, Jiang, Yi, Gao, Yiming, Ye, Hu, Du, Daniel~K, Yuan, Zehuan, \& Wu, Xinglong. 2025.
\newblock Tokenflow: Unified image tokenizer for multimodal understanding and generation.
\newblock {\em Pages  2545--2555 of:} {\em Proceedings of the Computer Vision and Pattern Recognition Conference}.

\bibitem[\protect\citename{Radford {\em et~al.}, }2021a]{radford2021learningclipsim}
Radford, Alec, Kim, Jong~Wook, Hallacy, Chris, Ramesh, Aditya, Goh, Gabriel, Agarwal, Sandhini, Sastry, Girish, Askell, Amanda, Mishkin, Pamela, Clark, Jack, {\em et~al.} 2021a.
\newblock Learning Transferable Visual Models from Natural Language Supervision.
\newblock {\em Pages  8748--8763 of:} {\em International Conference on Machine Learning}.
\newblock PMLR.

\bibitem[\protect\citename{Radford {\em et~al.}, }2021b]{Radford2021LearningTV}
Radford, Alec, Kim, Jong~Wook, Hallacy, Chris, Ramesh, Aditya, Goh, Gabriel, Agarwal, Sandhini, Sastry, Girish, Askell, Amanda, Mishkin, Pamela, Clark, Jack, Krueger, Gretchen, \& Sutskever, Ilya. 2021b.
\newblock Learning Transferable Visual Models From Natural Language Supervision.
\newblock {\em In:} {\em International Conference on Machine Learning}.

\bibitem[\protect\citename{Rombach {\em et~al.}, }2022]{Rombach2021HighResolutionIS}
Rombach, Robin, Blattmann, A., Lorenz, Dominik, Esser, Patrick, \& Ommer, Bj{\"o}rn. 2022.
\newblock High-Resolution Image Synthesis with Latent Diffusion Models.
\newblock {\em 2022 IEEE/CVF Conference on Computer Vision and Pattern Recognition (CVPR)}.

\bibitem[\protect\citename{Schuhmann \& Beaumont, }2022]{schuhmann2022laion}
Schuhmann, Christoph, \& Beaumont, Romain. 2022.
\newblock Laion-aesthetics.
\newblock {\em LAION. AI}.

\bibitem[\protect\citename{Somepalli {\em et~al.}, }2024]{csd}
Somepalli, Gowthami, Gupta, Anubhav, Gupta, Kamal, Palta, Shramay, Goldblum, Micah, Geiping, Jonas, Shrivastava, Abhinav, \& Goldstein, Tom. 2024.
\newblock Measuring Style Similarity in Diffusion Models.
\newblock {\em arXiv preprint arXiv:2404.01292}.

\bibitem[\protect\citename{Song {\em et~al.}, }2025]{song2025omniconsistency}
Song, Yiren, Liu, Cheng, \& Shou, Mike~Zheng. 2025.
\newblock Omniconsistency: Learning style-agnostic consistency from paired stylization data.
\newblock {\em arXiv preprint arXiv:2505.18445}.

\bibitem[\protect\citename{Su {\em et~al.}, }2024]{rope}
Su, Jianlin, Ahmed, Murtadha, Lu, Yu, Pan, Shengfeng, Bo, Wen, \& Liu, Yunfeng. 2024.
\newblock Roformer: Enhanced transformer with rotary position embedding.
\newblock {\em Neurocomputing}, {\bf 568}, 127063.

\bibitem[\protect\citename{Tan {\em et~al.}, }2024]{tan2024ominicontrol}
Tan, Zhenxiong, Liu, Songhua, Yang, Xingyi, Xue, Qiaochu, \& Wang, Xinchao. 2024.
\newblock Ominicontrol: Minimal and universal control for diffusion transformer.
\newblock {\em arXiv preprint arXiv:2411.15098}, {\bf 3}.

\bibitem[\protect\citename{Team, }2025]{decart2025lucyedit}
Team, DecartAI. 2025.
\newblock Lucy Edit: Open-Weight Text-Guided Video Editing.

\bibitem[\protect\citename{Wan {\em et~al.}, }2025]{wan2025}
Wan, Team, Wang, Ang, Ai, Baole, Wen, Bin, Mao, Chaojie, Xie, Chen-Wei, Chen, Di, Yu, Feiwu, Zhao, Haiming, Yang, Jianxiao, Zeng, Jianyuan, Wang, Jiayu, Zhang, Jingfeng, Zhou, Jingren, Wang, Jinkai, Chen, Jixuan, Zhu, Kai, Zhao, Kang, Yan, Keyu, Huang, Lianghua, Feng, Mengyang, Zhang, Ningyi, Li, Pandeng, Wu, Pingyu, Chu, Ruihang, Feng, Ruili, Zhang, Shiwei, Sun, Siyang, Fang, Tao, Wang, Tianxing, Gui, Tianyi, Weng, Tingyu, Shen, Tong, Lin, Wei, Wang, Wei, Wang, Wei, Zhou, Wenmeng, Wang, Wente, Shen, Wenting, Yu, Wenyuan, Shi, Xianzhong, Huang, Xiaoming, Xu, Xin, Kou, Yan, Lv, Yangyu, Li, Yifei, Liu, Yijing, Wang, Yiming, Zhang, Yingya, Huang, Yitong, Li, Yong, Wu, You, Liu, Yu, Pan, Yulin, Zheng, Yun, Hong, Yuntao, Shi, Yupeng, Feng, Yutong, Jiang, Zeyinzi, Han, Zhen, Wu, Zhi-Fan, \& Liu, Ziyu. 2025.
\newblock Wan: Open and Advanced Large-Scale Video Generative Models.
\newblock {\em arXiv preprint arXiv:2503.20314}.

\bibitem[\protect\citename{Wang {\em et~al.}, }2024]{wang2024instantstyle}
Wang, Haofan, Spinelli, Matteo, Wang, Qixun, Bai, Xu, Qin, Zekui, \& Chen, Anthony. 2024.
\newblock Instantstyle: Free lunch towards style-preserving in text-to-image generation.
\newblock {\em arXiv preprint arXiv:2404.02733}.

\bibitem[\protect\citename{Wang {\em et~al.}, }2025]{wang2025omnistyle}
Wang, Ye, Liu, Ruiqi, Lin, Jiang, Liu, Fei, Yi, Zili, Wang, Yilin, \& Ma, Rui. 2025.
\newblock OmniStyle: Filtering High Quality Style Transfer Data at Scale.
\newblock {\em Pages  7847--7856 of:} {\em Proceedings of the Computer Vision and Pattern Recognition Conference}.

\bibitem[\protect\citename{Wang {\em et~al.}, }2023a]{wang2023stylediffusion}
Wang, Zhizhong, Zhao, Lei, \& Xing, Wei. 2023a.
\newblock Stylediffusion: Controllable disentangled style transfer via diffusion models.
\newblock {\em Pages  7677--7689 of:} {\em Proceedings of the IEEE/CVF international conference on computer vision}.

\bibitem[\protect\citename{Wang {\em et~al.}, }2023b]{wang2023styleadapter}
Wang, Zhouxia, Wang, Xintao, Xie, Liangbin, Qi, Zhongang, Shan, Ying, Wang, Wenping, \& Luo, Ping. 2023b.
\newblock StyleAdapter: A Unified Stylized Image Generation Model.
\newblock {\em arXiv preprint arXiv:2309.01770}.

\bibitem[\protect\citename{Wu {\em et~al.}, }2025a]{wu2025qwen}
Wu, Chenfei, Li, Jiahao, Zhou, Jingren, Lin, Junyang, Gao, Kaiyuan, Yan, Kun, Yin, Sheng-ming, Bai, Shuai, Xu, Xiao, Chen, Yilei, {\em et~al.} 2025a.
\newblock Qwen-image technical report.
\newblock {\em arXiv preprint arXiv:2508.02324}.

\bibitem[\protect\citename{Wu {\em et~al.}, }2025b]{wu2025omnigen2}
Wu, Chenyuan, Zheng, Pengfei, Yan, Ruiran, Xiao, Shitao, Luo, Xin, Wang, Yueze, Li, Wanli, Jiang, Xiyan, Liu, Yexin, Zhou, Junjie, {\em et~al.} 2025b.
\newblock OmniGen2: Exploration to Advanced Multimodal Generation.
\newblock {\em arXiv preprint arXiv:2506.18871}.

\bibitem[\protect\citename{Xing {\em et~al.}, }2024]{xing2024csgo}
Xing, Peng, Wang, Haofan, Sun, Yanpeng, Wang, Qixun, Bai, Xu, Ai, Hao, Huang, Renyuan, \& Li, Zechao. 2024.
\newblock Csgo: Content-style composition in text-to-image generation.
\newblock {\em arXiv preprint arXiv:2408.16766}.

\bibitem[\protect\citename{Ye {\em et~al.}, }2025]{ye2025stylemaster}
Ye, Zixuan, Huang, Huijuan, Wang, Xintao, Wan, Pengfei, Zhang, Di, \& Luo, Wenhan. 2025.
\newblock Stylemaster: Stylize your video with artistic generation and translation.
\newblock {\em Pages  2630--2640 of:} {\em Proceedings of the Computer Vision and Pattern Recognition Conference}.

\bibitem[\protect\citename{Zhang {\em et~al.}, }2023a]{zhang2023adding}
Zhang, Lvmin, Rao, Anyi, \& Agrawala, Maneesh. 2023a.
\newblock Adding conditional control to text-to-image diffusion models.
\newblock {\em In:} {\em Proceedings of the IEEE/CVF international conference on computer vision}.

\bibitem[\protect\citename{Zhang {\em et~al.}, }2023b]{zhang2023forgedit}
Zhang, Shiwen, Xiao, Shuai, \& Huang, Weilin. 2023b.
\newblock Forgedit: Text guided image editing via learning and forgetting.
\newblock {\em arXiv preprint arXiv:2309.10556}.

\bibitem[\protect\citename{Zhang {\em et~al.}, }2025a]{zhang2025cdst}
Zhang, Shiwen, Chen, Zhuowei, Chen, Lang, \& Wu, Yanze. 2025a.
\newblock CDST: Color Disentangled Style Transfer for Universal Style Reference Customization.
\newblock {\em arXiv preprint arXiv:2506.13770}.

\bibitem[\protect\citename{Zhang {\em et~al.}, }2026]{zhang2026qwenstyle}
Zhang, Shiwen, Huang, Haibin, Zhang, Chi, \& Li, Xuelong. 2026.
\newblock QwenStyle: Content-Preserving Style Transfer with Qwen-Image-Edit.
\newblock {\em arXiv preprint arXiv:2601.06202}.

\bibitem[\protect\citename{Zhang {\em et~al.}, }2025b]{zhang2025easycontrol}
Zhang, Yuxuan, Yuan, Yirui, Song, Yiren, Wang, Haofan, \& Liu, Jiaming. 2025b.
\newblock Easycontrol: Adding efficient and flexible control for diffusion transformer.
\newblock {\em Pages  19513--19524 of:} {\em Proceedings of the IEEE/CVF International Conference on Computer Vision}.

\bibitem[\protect\citename{Zi {\em et~al.}, }2025]{zi2025senorita}
Zi, Bojia, Ruan, Penghui, Chen, Marco, Qi, Xianbiao, Hao, Shaozhe, Zhao, Shihao, Huang, Youze, Liang, Bin, Xiao, Rong, \& Wong, Kam-Fai. 2025.
\newblock Se$\backslash$\~{} norita-2M: A High-Quality Instruction-based Dataset for General Video Editing by Video Specialists.
\newblock {\em arXiv preprint arXiv:2502.06734}.

\end{thebibliography}
\bibliographystyle{authordate1}

\end{document}